\documentclass[10pt,twocolumn,letterpaper]{article}

\usepackage{cvpr}              

\definecolor{cvprblue}{rgb}{0.21,0.49,0.74}
\usepackage[pagebackref,breaklinks,colorlinks,allcolors=cvprblue]{hyperref}
\usepackage{multirow}
\def\paperID{*****} 
\def\confName{CVPR}
\def\confYear{2025}

\title{HCQA-1.5 @ Ego4D EgoSchema Challenge 2025}

\author{
Haoyu Zhang$^{1\,2}$, Yisen Feng$^{1}$, Qiaohui Chu$^{1\,2}$, Meng Liu$^{3}$, Weili Guan$^{1}$, \\ Yaowei Wang$^{1\,2}$, Liqiang Nie$^{1}$\\
$^1$Harbin Institute of Technology (Shenzhen) \qquad  $^2$Pengcheng Laboratory    \\$^3$Shandong Jianzhu University\\
{\tt\small \{zhang.hy.2019, yisenfeng.hit, qiaohuichu8599, mengliu.sdu, honeyguan, nieliqiang\}@gmail.com;} \\ {\tt\small wangyw@pcl.ac.cn}
}

\begin{document}
\maketitle
\begin{abstract}
In this report, we present the method that achieves third place for Ego4D EgoSchema Challenge in CVPR 2025.
To improve the reliability of answer prediction in egocentric video question answering, we propose an effective extension to the previously proposed HCQA framework. Our approach introduces a multi-source aggregation strategy to generate diverse predictions, followed by a confidence-based filtering mechanism that selects high-confidence answers directly. For low-confidence cases, we incorporate a fine-grained reasoning module that performs additional visual and contextual analysis to refine the predictions. Evaluated on the EgoSchema blind test set, our method achieves 77\% accuracy on over 5,000 human-curated multiple-choice questions, outperforming last year’s winning solution and the majority of participating teams. 
Our code will be added at \href{https://github.com/Hyu-Zhang/HCQA}{https://github.com/Hyu-Zhang/HCQA}.
\end{abstract}    
\section{Introduction}

Egocentric video understanding has become a key research focus in the fields of embodied AI~\cite{pmlr-v235-zhang24aj} and video-language modeling~\cite{zhang2023attribute,zhang2021multimodal,guan2022bi}, offering valuable insights into how people interact with the world from a first-person perspective. Compared to third-person video, egocentric footage captures fine-grained, context-rich visual signals, but it also poses significant challenges~\cite{feng2025object,feng2024objectnlq}. These include rapid camera motion, limited field of view, and frequent occlusions, all of which complicate downstream reasoning tasks~\cite{zhang2025exo2ego}.

One such representative task is the EgoSchema~\cite{mangalam2023egoschema} Challenge from the Ego4D benchmark~\cite{grauman2022ego4d}, which focuses on long-form video question answering. The goal is to select the correct answer from five multiple-choice options, given a three-minute-long egocentric video and an associated question. Evaluation is conducted on the EgoSchema dataset~\cite{mangalam2024egoschema}, which includes over 5,000 human-curated question–answer pairs spanning more than 250 hours of real-world egocentric video. The dataset covers a wide range of natural human activities and presents a particularly challenging benchmark due to its long temporal context and semantic ambiguity.

While recent progress in large language models (LLMs) and vision-language models (VLMs) has advanced performance on video question answering~\cite{zhang2023uncovering,wang2025time}, existing methods such as HCQA~\cite{zhang2024hcqa}, often rely on a single LLM in the final decision stage. This can lead to overconfidence in incorrect predictions, especially when visual cues are sparse or ambiguous. To address this, we propose a simple yet effective extension to the HCQA framework that improves answer reliability through two key enhancements: (1) multi-source aggregation, where multiple advanced LLMs are used to produce diverse candidate answers; and (2) fine-grained reasoning, where low-confidence outputs are selectively reprocessed using visual and textual reasoning modules. This two-stage strategy enables more robust decision-making in challenging scenarios without introducing significant architectural complexity.

By adopting this enhanced framework, our approach outperforms last year’s winning solution as well as the majority of participating teams in the current challenge, ultimately securing third place in the final leaderboard. Specifically, our pipeline achieves a notable improvement over HCQA, increasing accuracy from 75\% to 77\%, as shown in Table~\ref{tab:leaderboard}.

\section{Methodology}

\begin{figure*}[t]
  \centering
   \includegraphics[width=0.8\linewidth]{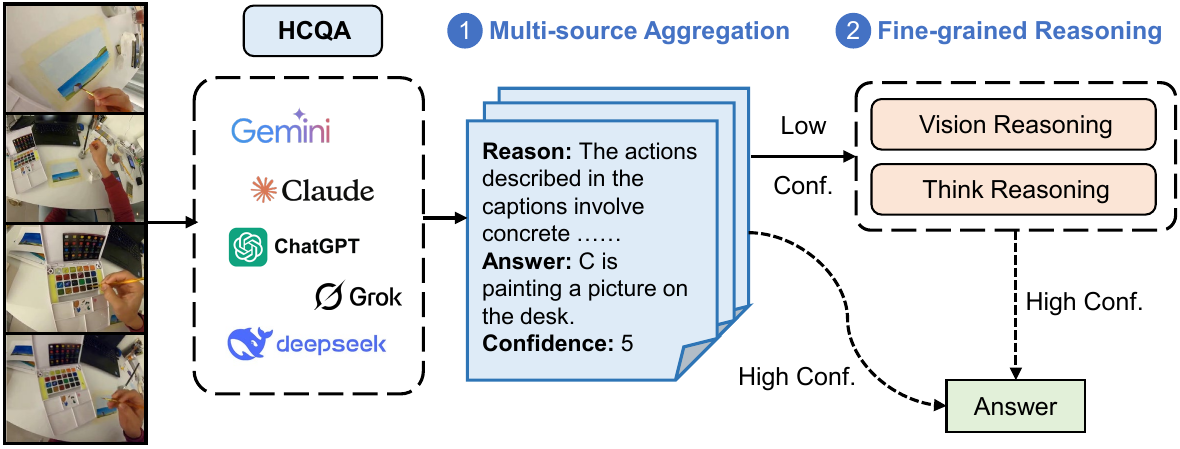}
   \caption{An illustration of two-stage decision-making process.}
   \label{fig:method}
\end{figure*}

As illustrated in Figure~\ref{fig:method}, given a long egocentric video, we first apply the previously proposed HCQA framework~\cite{zhang2024hcqa} to obtain multiple diverse predictions. These initial outputs serve as the basis for a two-stage decision-making process aimed at improving the reliability of the final answer. In the first stage, we perform multi-source aggregation to consolidate the results. High-confidence answers are directly selected as final outputs, while low-confidence ones are forwarded to a fine-grained reasoning module for further refinement.

\subsection{Multi-source Aggregation}

To enhance prediction diversity, we replace the third-stage LLM used in the original HCQA pipeline, such as GPT-4o, with several state-of-the-art models, including Gemini-1.5-Pro\footnote{\url{https://deepmind.google/models/gemini/pro/}.}, GPT-4.1\footnote{\url{https://openai.com/index/gpt-4-1/}.}, and Qwen2.5~\cite{yang2025qwen3}. The use of multiple LLMs introduces complementary perspectives, which improves robustness and coverage across different input cases. To effectively integrate the outputs, we adopt a confidence-based filtering mechanism. For each sample, predictions with a confidence score higher than 4 (on a scale from 1 to 5) are considered reliable and retained as final answers. The remaining low-confidence predictions are passed to the next stage for additional processing.

\subsection{Fine-grained Reasoning}

For samples with low-confidence predictions, we introduce a fine-grained reasoning module that includes two complementary strategies: vision-based reasoning and thought-based reasoning. The vision-based approach focuses on re-analyzing visual evidence. Specifically, we extract 45 frames uniformly from the original video and input them into the Qwen2.5-VL-72B model~\cite{bai2025qwen2}  with default settings to generate refined predictions based on visual content.

In contrast, the thought-based approach emphasizes textual and contextual reasoning. It aggregates all available textual information, including captions and summaries extracted by HCQA, as well as all predictions generated in the multi-source aggregation stage. This information is then fed into DeepSeek-R1~\cite{guo2025deepseek}, which performs deeper thinking reasoning to produce an updated answer.

Finally, we compare the outputs from both reasoning strategies and select the one with the higher confidence score as the final result. This combined approach allows the system to recover reliable answers even from initially uncertain predictions, improving the overall robustness and accuracy of the framework.
\section{Experiment}

\begin{table}
  \caption{Performance comparison of existing work and the top five teams on the public leaderboard.}
  \centering
  \begin{tabular}{ccc}
    \toprule
    \textbf{Method} & \textbf{Rank} & \textbf{Accuracy} \\
    \midrule
    mPLUG-Owl~\cite{ye2023mplug} & -&0.31\\
    LongViViT~\cite{papalampidi2023simple}& -&0.33\\
    InternVideo2~\cite{wang2024internvideo2} &-& 0.41\\
    LLoVi~\cite{zhang2023simple} &-& 0.50\\
    VideoAgent~\cite{wang2024videoagent} & -&0.54\\
    ProViQ~\cite{choudhury2023zero}&-&0.57\\
    LifelongMemory~\cite{wang2023lifelongmemory}&-&0.68\\
    Gemini-1.5-Pro~\cite{reid2024gemini} & -&0.71\\
    GPT-4o~\cite{hurst2024gpt}&-&0.72\\
    Qwen2.5-VL-72B~\cite{bai2025qwen2}&-&0.76\\
    \midrule
    Noah's\_Ark\_Lab & 5 & 0.75\\
    ccego & 4 & 0.76\\
    L\_PCIE (PCIE) & 2& 0.79\\
    Reality Distortion & 1 & 0.81\\
    \midrule
   HCQA-1.5 (iLearn2.0) & 3& \textbf{0.77}\\
    \bottomrule
  \end{tabular}
  \label{tab:leaderboard}
\end{table}

\begin{figure*}[t]
  \centering
   \includegraphics[width=\linewidth]{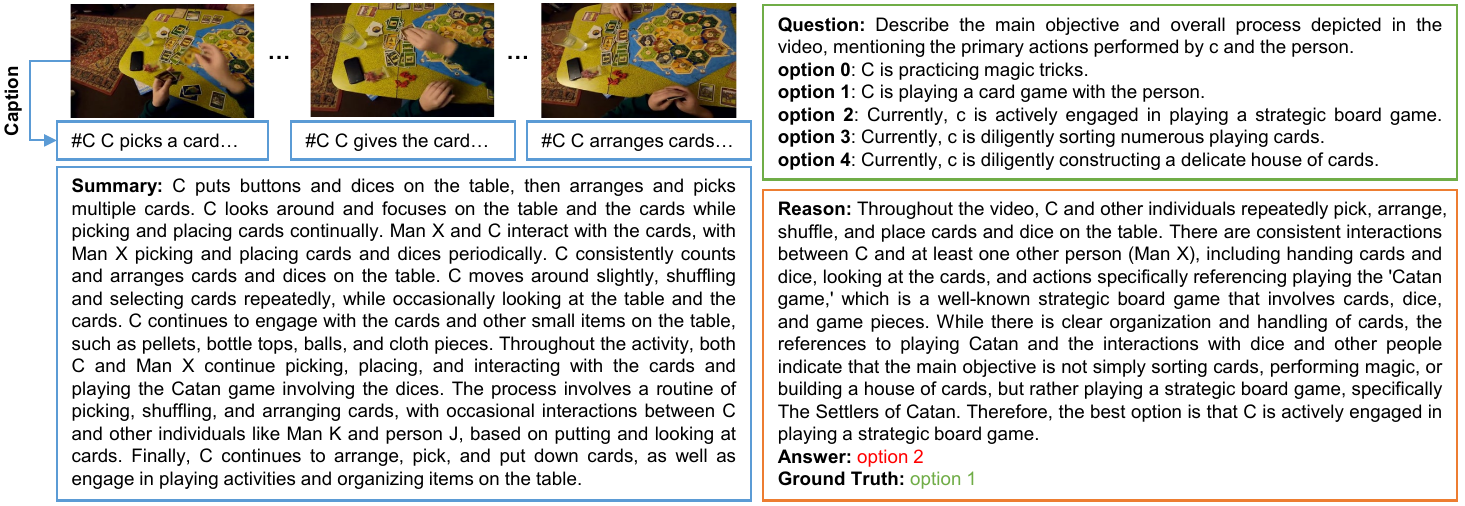}
   \caption{One failed example of our framework on EgoSchema subset.}
   \label{fig:case}
\end{figure*}
\subsection{Performance Comparison}
Table~\ref{tab:leaderboard} presents the performance comparison of existing methods and the top five teams on the public leaderboard of the Ego4D EgoSchema Challenge. Our method, HCQA-1.5 (iLearn2.0), achieves an accuracy of 77\%, securing third place among all submissions. It outperforms several strong baselines, including GPT-4o (72\%) and Gemini-1.5-Pro (71\%), as well as other top teams such as Noah’s\_Ark\_Lab (75\%) and ccego (76\%). The results demonstrate the effectiveness of our multi-source aggregation and fine-grained reasoning in handling complex, long-form egocentric video question answering.

\subsection{Ablation Study}
Table~\ref{tab:ablation} presents an ablation study assessing the contributions of different models at each stage of our HCQA-1.5 framework. In Stage 1, the individual models Gemini-1.5-Pro, Qwen2.5, and GPT-4.1 yield varying levels of accuracy, with GPT-4.1 achieving the highest at 76.1\%. Integrating their outputs by selecting the answer with the highest confidence leads to further performance gains, demonstrating the advantage of model ensemble.

In Stage 2, Qwen2.5-VL-72B and DeepSeek-R1 perform similarly (76.8\% vs. 76.6\%) when applied to low-confidence samples, confirming the value of fine-grained reasoning in uncertain cases.

Combining both stages into the full pipeline results in the highest overall accuracy of 77.3\%, indicating that each stage makes a positive contribution to the final performance.

\begin{table}[t]
  \caption{Ablation study with different models of our framework.}
  \centering
  \begin{tabular}{ccccc}
    \toprule
    \textbf{Stage} & \textbf{Model} & \textbf{Accuracy} \\
    \midrule
    \multirow{4}{*}{1}&Gemini-1.5-Pro&0.710\\
    &GPT-4.1&0.761\\
    &Qwen2.5&0.748\\
    &Integration&0.763\\
    \midrule
    \multirow{2}{*}{2}&Qwen2.5-VL-72B&0.768\\
    &DeepSeek-R1&0.766\\
    \midrule
    -&HCQA-1.5&\textbf{0.773}\\
    
    \bottomrule
  \end{tabular}
  \label{tab:ablation}
  \vspace{-1ex}
\end{table}

\subsection{Case Study}
Figure~\ref{fig:case} shows a failed example of our framework. The model incorrectly selects option 2 (“C is actively engaged in playing a strategic board game”) instead of the correct option 1 (“C is playing a card game with the person”). This mistake is mainly due to the model's overreliance on surface-level visual cues—such as the presence of dice and cards—that are commonly associated with strategic board games like ``The Settlers of Catan''. However, the question emphasizes not just the objects but the overall objective and interaction between individuals.

While the summary clearly describes C and another person repeatedly picking, handing, and arranging cards together, the model fails to capture this collaborative aspect and instead focuses on categorizing the scene based on game components. This suggests a weakness in understanding social dynamics and intent, particularly when multiple plausible interpretations share similar visual elements.

\section{Conclusion}
We propose an effective enhancement to the HCQA framework for egocentric video question answering, integrating multi-source aggregation with fine-grained reasoning. By selectively reasoning over low-confidence cases, our method improves the reliability of final answers. Achieving 77\% accuracy on the EgoSchema benchmark and ranking third in the CVPR 2025 Challenge, our approach demonstrates strong performance of our design.

{
    \small
    \bibliographystyle{ieeenat_fullname}
    \bibliography{main}
}


\end{document}